\documentclass[letterpaper]{article} 
\usepackage{aaai2026}  
\usepackage{times}  
\usepackage{helvet}  
\usepackage{courier}  
\usepackage[hyphens]{url}  
\usepackage{graphicx} 
\usepackage{natbib}  
\usepackage{caption} 
\usepackage{algorithm}
\usepackage{algpseudocode}
\usepackage{amsmath}
\usepackage{amsfonts}
\usepackage{multirow}
\usepackage{nameref}
\newcommand{\mathbbm}[1]{\text{\usefont{U}{bbm}{m}{n}#1}} 

\usepackage{newfloat}
\usepackage{listings}
\DeclareCaptionStyle{ruled}{labelfont=normalfont,labelsep=colon,strut=off} 
\floatstyle{ruled}
\newfloat{listing}{tb}{lst}{}
\floatname{listing}{Listing}
\title{Fractured Glass, Failing Cameras: Simulating Physics-Based Adversarial Samples for Autonomous Driving Systems}
\author{
    Manav Prabhakar, Jwalandhar Girnar, Arpan Kusari\thanks{Corresponding author} 
}
\affiliations{
    University of Michigan Transportation Research Institute\\
    2901 Baxter Road, Room 202, \\
    Ann Arbor, MI-48103\\
    \{prmanav, jwala, kusari\}@umich.edu
}

\usepackage{bibentry}

\begin{document}

\maketitle

\begin{abstract}
While much research has recently focused on generating physics-based adversarial samples, a critical yet often overlooked category originates from physical failures within on-board cameras—components essential to the perception systems of autonomous vehicles. Camera failures, whether due to external stresses causing hardware breakdown or internal component faults, can directly jeopardize the safety and reliability of autonomous driving systems. Firstly, we motivate the study using two separate real-world experiments to showcase that indeed glass failures would cause the detection based neural network models to fail. Secondly, we develop a simulation-based study using the physical process of the glass breakage to create perturbed scenarios, representing a realistic class of physics-based adversarial samples. Using a finite element model (FEM)-based approach, we generate surface cracks on the camera image by applying a stress field defined by particles within a triangular mesh. Lastly, we use physically-based rendering (PBR) techniques to provide realistic visualizations of these physically plausible fractures. To assess the safety implications, we apply the simulated broken glass effects as image filters to two autonomous driving datasets- KITTI and BDD100K- as well as the large-scale image detection dataset MS-COCO. We then evaluate detection failure rates for critical object classes using CNN-based object detection models (YOLOv8 and Faster R-CNN) and a transformer-based architecture with Pyramid Vision Transformers. To further investigate the distributional impact of these visual distortions, we compute the Kullback-Leibler (K-L) divergence between three distinct data distributions, applying various broken glass filters to a custom dataset (captured through a cracked windshield), as well as the KITTI and Kaggle cats and dogs datasets. The K-L divergence analysis suggests that these broken glass filters do not introduce significant distributional shifts. Our goal is to provide a robust, physics-based methodology for generating adversarial samples that reflect real-world camera failures, with the overarching aim of improving the resilience and safety of autonomous driving systems against such physical threats.
\end{abstract}

\begin{links}
    \link{Code}{https://github.com/manavprabhakar/camera-failure}
\end{links}

\section{Introduction}
\label{sec:intro}
\begin{figure*}[btp]
    \centering
    \includegraphics[width=0.7\linewidth]{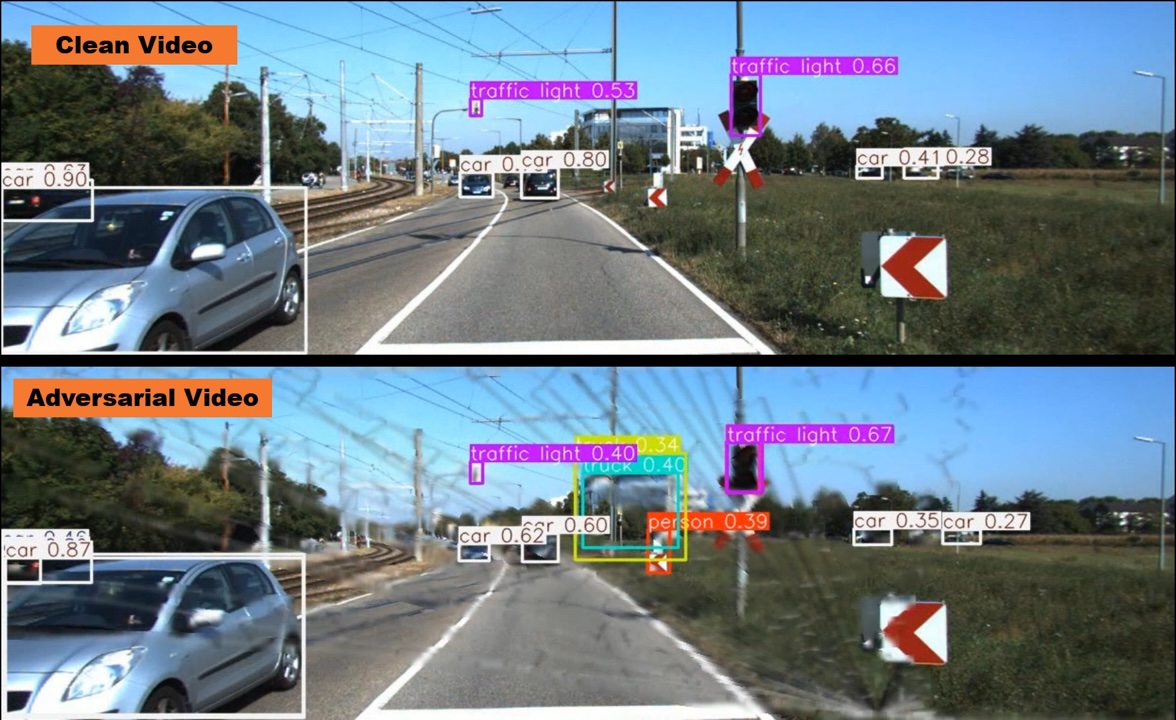}
    \caption{A qualitative comparison of clean vs adversarial video generated using our simulation and rendering method on KITTI. This frame shows false positives, and reduced confidence levels for true positives. Refer to the supplementary material for full video.}
    \label{fig:overview}
\end{figure*}
Cameras are ubiquitous as remote sensors, collecting data from an unstructured and dynamic external environment, often in harsh conditions. A failure or fault in a sensor is a divergence from the functional state in at least one given parameter of the system \citep{van1997remarks}. These faults can occur due to internal (such as wear and tear) or external (temperature, humidity etc) causes. For RGB cameras, internal causes include dead pixels while external causes include fractured enclosures or outer lens, and condensation. These abrupt failures are hard to detect and negatively impact object detection algorithms - reducing accuracy and often leading to hallucination as shown in Fig. \ref{fig:overview}. The failures occurring in an automated vehicle (AV) for example, can lead to critical safety issues resulting in crashes and in some cases, fatalities. 

Currently, to the best of authors' knowledge, there are no rigorous methods for generating camera based sensor failures \citep{ceccarelli2022rgb}.

In this work, we focus on the sensor failure occurring due to fractures in any glass covering a camera (or camera enclosure), although the process detailed in this paper can be used for any of the camera failures listed in \citep{ceccarelli2022rgb}. These glass fracture effects in a camera can be caused due to an external object hitting the camera or as a result of heat and/or pressure developing suddenly within the enclosure. In the parlance of neural networks, an image captured in such conditions is considered as an adversarial sample. 
Previous research \citep{akhtar2018threat, carlini2017adversarial, szegedy2013intriguing} shows that even small amounts of corruptions, sometimes difficult to be seen by human eyes, are enough to completely fool the neural networks where a subtle change of inputs can lead to a drastic change in outputs. We would like to note that \cite{li2019adversarial} provided a physical camera-based adversarial attack paradigm, which serves as the closest related work in this domain. They presented a modification of the image using an overlay of a translucent, carefully crafted sticker which led to misclassification. 

\begin{figure*}
    \centering
    \includegraphics[width=0.8\textwidth]{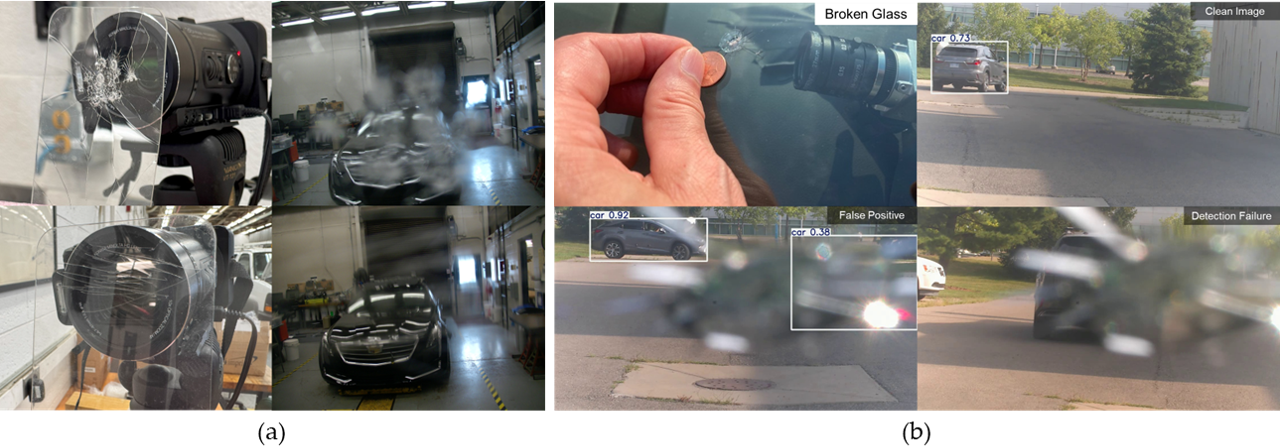}
    \caption{(a) Indoor static experiment. Left: Camera with 2 different fractured tempered glass patterns; right - images of the vehicle under the different fractures. (b) Outdoor dynamic experiment. Top left - a coin sized windshield crack; top right - clean image with the vehicle detected using YOLOv8; bottom left - false positive through the crack; bottom right - detection failure through the glass. More examples from these experiments have been provided in the supplementary material.}
    \label{fig:real_broken_lens}
\end{figure*}

To understand the effect of these fractures on the resulting camera images, we conducted two distinct experiments: one in an indoor static environment and the other in a dynamic outdoor environment. The first one involved fracturing tempered glass and placing it in front of the camera (see Fig. \ref{fig:real_broken_lens}(a)) with a static vehicle in the scene to understand how different fracture patterns affect the quality and appearance of the scene. We captured images at different focal lengths to judge the variability of such corruptions. This helped us answer certain qualitative questions about the visual appearance of these fractures with respect to their spread and intensity, motivating our approach in Section \nameref{sec:focal plane}. The experimental setup and the detailed experimental results are in Sec. \nameref{sec:static} of Supplementary. 
The second experiment (Fig. \ref{fig:real_broken_lens}(b)) consisted of recording an outdoor video with dynamic vehicles under daylight conditions by placing a MobileEye camera next to a windshield crack presented in Fig. \ref{fig:real_broken_lens} (shown in the upper left) and performing inference using YOLOv8 \citep{JocherUltralyticsYOLO2023}) to gain a primitive understanding of the impact of such scenarios on object detection networks. We observed that the model can easily detect the vehicle in a clean image while it suffers from detection failure (lower right) or generates false positives (lower left). Interestingly, the presence of a crack can also unexpectedly increase the confidence in prediction of the car presenting a clearly defined edge (0.92 in the lower left vs. 0.75 in the upper left). The detailed inference results with vehicle and person class is given in Sec. \nameref{sec:dynamic} of Supplementary.

We then looked for real broken glass images online (Sec. \nameref{sec:real_glass_fracture} of Supplementary) but failed to build a dataset large enough to enable a data-driven approach for adversarial defense for these conditions. Additionally, we experimented with CGI tools like Maya and Blender for creating such effects but they lack the flexibility, control, scale and physics to simulate these conditions. The closest simulation option in existing literature is ArcSim \citep{pfaff2014adaptive}. However, their high-resolution simulation outputs are extremely slow ($\approx$ 20 hours), making it difficult to scale. As a result,we directed our efforts towards creating a scalable simulation-based pipeline for generating fractures that can be used to advance perception stack.   

 For a glass fracture, the principal point, force and angle of incidence may be random, but the spread and the resulting pattern follows an inherently physical process (being either linear or radial). We thus build a fracture simulation based on particles in a triangular mesh generated randomly and perform stress propagation through the mesh. Our simulation allows us to produce the fractures within a triangular mesh at every discrete time state $\delta t$. We use OpenCV to convert the given mesh to a corresponding broken glass pattern image. We then utilize physically-based rendering (PBR) \citep{pharr2023physically} to realistically render the surface fractures using bidirectional reflectance distribution function (BRDF) by calculating the amount of light reflected from a given point on a surface as a result of source(s) of light being incident on it. 

Combining our rendering approach with three popular open source datasets - KITTI \citep{geiger2013IJRR}, BDD100k \cite{yu2020bdd100k} and MS-COCO \citep{lin2014microsoft}, we are able to generate adversarial images efficiently. A common process for testing the generated adversarial images is to find the number of false positives/negatives across the image space. However, in our case, due to the adversarial effect being local, we cannot rely simply on an image based measure. We therefore, use the adversarial images (similar to the lower left figure of Fig. \ref{fig:real_broken_lens}) and extract the objects which lie within the region where the fracture exists using the ground truth bounding boxes. We then utilize YOLOv8,  Faster R-CNN \citep{ren2016faster} and Pyramid Vision Transformer (PVTv2) \cite{wang2022pvt} to find the percentage of objects that fail when the adversarial filters are applied. 
We also provide ablation studies to understand the distributional differences between the three set of images: Real broken glass images collected experimentally, real broken glass images collected online and the generated images. We compute the Kullbeck-Liebler (K-L) divergence for these image distributions to prove similarity of the generated images to the real broken glass images. We utilize cat images from Kaggle Cats and Dogs dataset as control to understand the difference between image distributions \citep{kaggle}.

\noindent
The major contributions of the paper can be summarized as follows:
\begin{itemize}
    \item We provide a novel way of abstracting glass fracture through a combination of stress propagation methods and minimum spanning trees, to generate physically sound broken glass patterns. 
    \item We present a PBR approach to facilitate a realistic render of camera failures that can be used with any kind of existing computer vision datasets - both images and videos. 
    \item Our simulation and rendering pipelines are scalable and computationally efficient ($\approx 1.6s$) allowing it to be used by both academia and industry for enhancing robust and out of distribution protection for a wide range of applications.
\end{itemize}
\section{Background}
\label{sec:background}
\subsection{Physics based adversarial samples}
The problem of adversarial sample can be defined as follows: for a model $M$ that classifies an input sample $X$ correctly to its designated class i.e. $M(X) = y_{true}$, adding an error $\epsilon$ to the input sample $X$, results in an altered sample $\hat{X}$ such that $M(\hat{X}) \neq y_{true}$. Thus, the injection of the error $\epsilon$ results in an adversarial sample that causes the model to fail. 

Although the idea of adversarial manipulation of the model has been identified in the context of machine learning quite some time ago \citep{dalvi2004adversarial}, in the last decade, the focus has squarely been on the adversarial attacks on neural networks \citep{szegedy2013intriguing, goodfellow2014explaining}. In these papers, the researchers showed that a small targeted injection of noise, almost imperceptible to the human eye, changed the labels completely \citep{szegedy2013intriguing} and conversely, images could be generated that looked completely unrecognizable to humans but which had perfect classifications from the DNNs \citep{nguyen2015deep}. 

While these adversarial samples probe the model for possible failures, they lack any physical realism behind their generation and need access to the model. To address this, some recent research has targeted building physically relevant adversarial samples. One of the first forays into this was made by \cite{kurakin2018adversarial} who targeted the accuracy of the models in the physical world by feeding noisy images from a cell-phone camera that led the model to incorrectly classify a large fraction of the samples. Along the similar vein, \cite{eykholt2018robust} demonstrated that real traffic signs can be perturbed with simple physical stickers placed strategically to fool state-of-the-art DL algorithms almost perfectly even with viewpoint changes. Other researchers have placed adversarial images \citep{kong2020physgan}, translucent patches on camera \citep{zolfi2021translucent} or artificial LiDAR surfaces \citep{tu2020physically} to generate samples which fool object detectors. While these prior research use physics in terms of generating the samples, they do not come from modeling a rigorous physical process and we aim to fill this gap in this work.

\subsection{Cracked/fractured glass theory}
The subject of how glass breaks and how it propagates is still an open research question and one that has been contentious with multiple physical theories being proposed \citep{rouxel2012flow}. While the microscopic procedure of glass crack is being debated on, on a macroscopic level, the cracking dynamics is well understood. \cite{liu2021analysis} analyzed the process of cracking of glass lens in the precision glass molding application using FEM with a three-dimensional model in a physical simulation software. The physical parameters were input into the software and the crack paths were analyzed using the simulation results. The authors performed a temperature and stress simulation of a high-precision three-dimensional mesh model of the molded glass. \citep{iben2009generating} provided a way to generate surface fractures in variety of materials including glass. As already mentioned in the introduction, \cite{pfaff2014adaptive} provided the simulation of  glass breaking as a thin sheet which forms the closest related work to our proposed method.

\section{Methodology}
\label{sec:method}
Generating realistic glass failures require creating large-scale physics based simulations by solving fracture dynamics on a triangulated finite element mesh with glass properties. 

\subsection{Broken glass simulation}

We represent glass using particles sampled from a uniform distribution spread across a plane constrained in the form of a 2D mesh using constrained Delaunay triangulation. This removes ill-shaped triangles and avoids uneven and unrealistic edges. 

Each particle $p_{i}$ has a position $x_{i}$ and has nearest neighbors $\mathbf{k_{i}}$ within a radius $r$ which have existing edges with $p_{i}$. Mathematically, the triangulation mesh $\mathcal{M}$ represents a finite set of 2-simplices such that if
\begin{equation}
\forall (K, K') \in \mathcal{M} \times \mathcal{M}, |K| \cap |K'| = |K \cap K'|.
\end{equation}
The crack patterns in glass occur due to stress from the external force (F) at the initial impact point $p_{I}$ by assuming a specific deformation law (elasticity and plasticity) of the glass (G) \citep{kuna2013finite}. We then compute the strength parameters in the form of effective stress $\sigma_{V}$ at the impact point (V) as the stress state of the impact point. The critical stress values for the strength of glass $\sigma_{C}$ is found using tests on simple samples with elementary loading conditions (e.g. tension test). The fracture then occurs when the effective stress is larger than the critical stress divided by the safety factor (S):
\begin{equation}
    \sigma_{V} (G, F)  > \dfrac{\sigma_{C}}{S}.
    \label{eq:safety-condition}
\end{equation}
From the classical theory of strength of materials, we know that the failure in most cases is controlled by the principal stresses $\sigma_{I}$ and $\sigma_{II}$ for 2D elements. The initial crack happens either by the normal-planar crack where the fracture faces are located perpendicular to the direction of the highest principal stress $\sigma_{I}$ \citep{rankine1857ii} or shear-planar crack where the fracture faces coincide with the intersection planes of the maximum shear stress $\tau_{max} = (\sigma_{I} - \sigma_{II})/2$ \citep{coulumb1776essai}. In the case of glass, we assume that the initial fracture happens perpendicular to the direction of the maximum principal stress.

From the initial impact point $p_{I}$, the stress propagation through glass is unstable since the crack grows abruptly without the need to increase external loading. 
From $p_{I}$, stress propagates in the vertex neighborhood $\mathbf{k_{i}}$ as the stress along the direction $\overrightarrow{p_{I}p_{j}}$ where $p_{j} \in \mathbf{k_{i}}$ as 
\begin{equation}
\sigma_{p_{j}} = \sigma_{V} * \dfrac{\overrightarrow{p_{I}p_{j}} \cdot \overrightarrow{n}}{|\overrightarrow{p_{I}p_{j}}||\overrightarrow{n}|}.
\end{equation}

\begin{figure}[!htbp]
    \centering
    \includegraphics[width=0.7\linewidth]{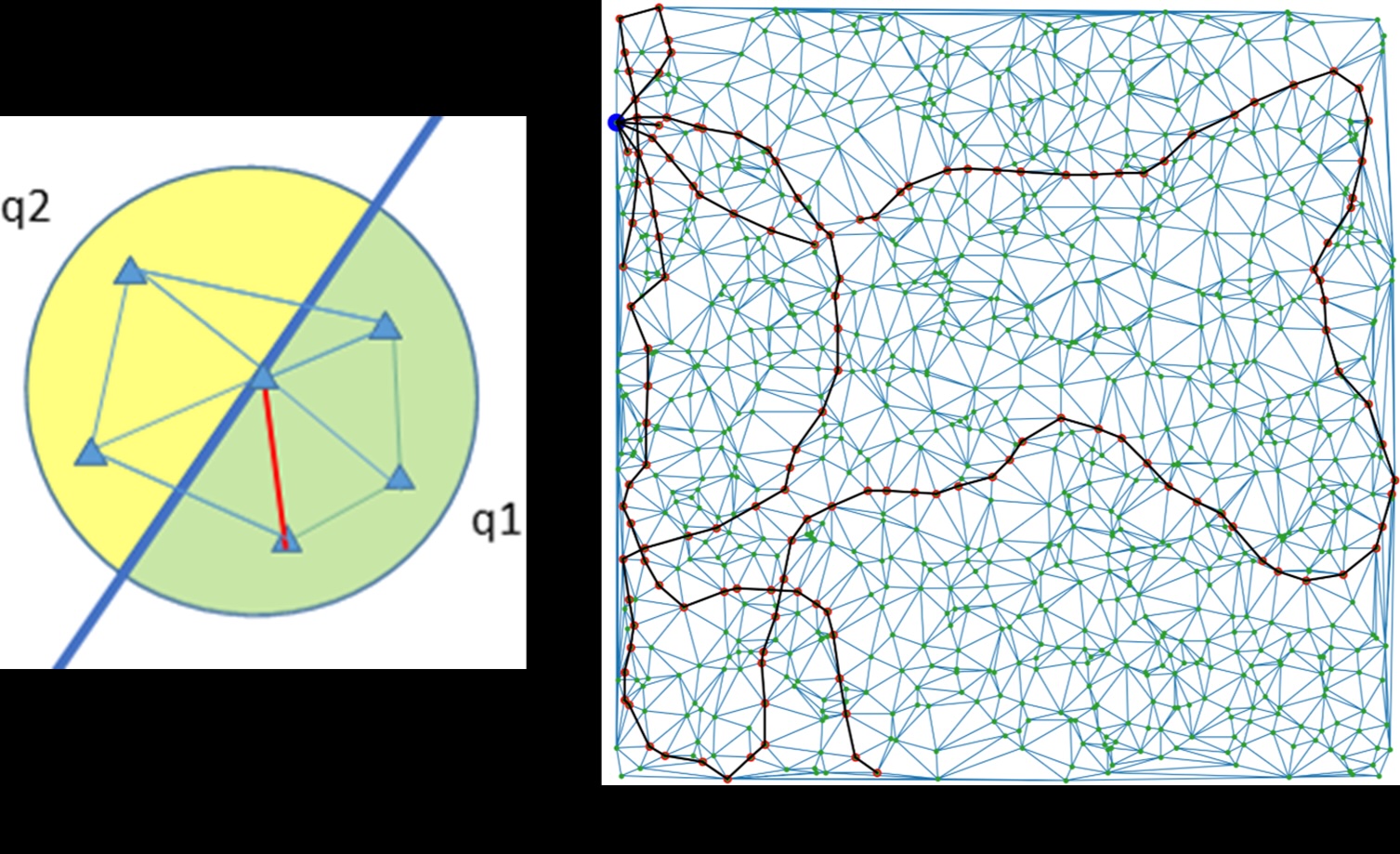}
    \caption{(a) For a splitting plane given in blue, the summed positive stress $q_1$ and summed negative stress $q_2$ are compared and the propagation happens on the side with greater summed stress and chosen node which is the closest to the splitting plane (given in red). (b) Shows how we simulate a fracture in a mesh originating from its impact point (marked in blue) to the nodes experiencing stress beyond their threshold strength (marked in red).}
    \label{fig:splitting-plane}
\end{figure}

With the stress calculated for each edge, the summed positive stress (showed in Fig. \ref{fig:splitting-plane}) can then be given as:
\begin{equation}
q_1 = \int_{\partial \Omega} \sigma_{p_j}\mathbbm{I}(\sigma_{p_j} > 0) dA
\end{equation}
for continuous surface $\Omega$ where $\mathbbm{I}$ is the indicator function. The summed positive stress for discrete simplices in the corresponding area A of radius R is given as
\begin{equation}
    q_1 = \sum_{K \in {A_R}} \sigma_{K}\mathbbm{I}(\sigma_{K} > 0).
\end{equation}
Similarly, the summed negative stress $q_2$ is calculated. Then for greater magnitude $\max(|q_1|, |q_2|)$, we choose the corresponding edge with the highest concentration of stress in the given segment as the optimal splitting plane since that provides the maximum stress relief. Thus, the stress travel along the mesh edges, dissipating the stress at each node point. 

The recursive application of the stress propagation is run until convergence of the stress in all states i.e. $\sigma_{p}^{(t)} 
\simeq \sigma_{p}^{(t-1)} \forall p \in V$. 

Propagating the stress in all directions across all nodes, results in back-cracking as explained in \cite{Obrien:1999:GMA}. To avoid it, we propagate only along the edges where the stress levels are maximum but perform a stress update on all neighboring nodes. We then use a minimum spanning tree (MST) on a mesh created using these stressed nodes. We combine this MST with our initial stress propagation field along the edges to compute the final crack pattern. The MST is an effective abstraction because it connects the nodes which are closer to each other and within the high stress field while removing redundancies.

Our computational process of stress propagation is defined in Algorithm \ref{alg:sim_alg} in Supplementary.

\subsection{Physically-based rendering}

Once we have generated the fractures at the mesh-level, our next goal is to create a visual render of these fractures. Like all PBR techniques, our method is based on the microfacet theory which states that any surface can be described by tiny little perfectly reflective mirrors called microfacets \citep{pharr2023physically}.

In accordance with the microfacet theory and energy conservation, we use the reflectance equation,
\begin{equation}
    L_{o}(x,{\omega}_{o},\lambda,t) = L_{e}(x,\omega_{o},\lambda,t) + L_{r}(x,\omega_{o},\lambda,t)
\end{equation}
where $L_{o}(x,{\omega}_{o},\lambda,t)$ is the total spectral radiance of wavelength $\lambda$ directed outward along direction $\omega_{o}$ at time t, from a particular position $x$.
$\omega_{o}$ is the direction of the outgoing light.
$t$ is time. $L_{e}$ is the emitted spectral radiance and $L_{r}$ is the reflected spectral radiance. 

\begin{figure*}[!ht]
    \centering
    \includegraphics[width=\textwidth]{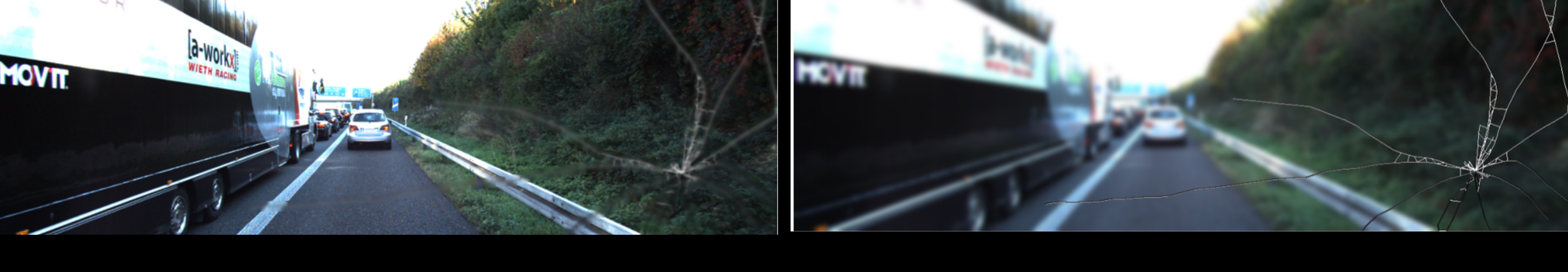}
    \caption{(a) Shows the simulated image with the road and vehicles in the focal plane (PBR and Far-focus). (b) denotes the simulated crack pattern in the focal plane (PBR and short focus).}
    \label{fig:PBR}
\end{figure*}

Let $I_1$ be the bidirectional reflectance distribution function,
\begin{equation*}
    I_1 = f_r(x,\omega_{i},\omega_{o},\lambda,t)
\end{equation*}
and let $I_2$ be the spectral radiance coming inward towards x from direction $\omega_i$ at time t. 
\begin{equation*}
    I_2 = L_i(x,\omega_i,\lambda,t)
\end{equation*}
Then, $L_r$ can be defined as 
\begin{equation}
    L_{r}(x,\omega_{o},\lambda,t) = \int_{\Omega} I_1 \cdot I_2 \cdot (\omega_i\cdot \textbf{n}) \,d\omega_i
\end{equation}
where $\Omega$ is the unit hemisphere centered around surface normal $\textbf{n}$  over $\omega_i$ such that $\omega_i \cdot n > 0$.

Abstracting the reflectance equation, we aim to create a visual render of our broken glass mesh. We have $L_e = 0$ as glass does not emit light. Now for calculating $L_r$, we consider any crack between the nodes as a microfacet. 
Then, we can define $L_r$ for every crack as:
\begin{equation}
    L_r = L_i \: (\omega_i \cdot \hat{\textbf{n}})
\end{equation}
Given the unit vectors ($\hat{\omega}_{\alpha}$) and  ($\hat{\omega}_{\theta}$) corresponding to the azimuth ($\alpha$) and zenith ($\theta$) angles respectively, we compute the mean energy incident on the crack as 
\begin{equation}
    \mathbb{E}(L_r) = \frac{|\hat{\omega}_{\alpha}\cdot\hat{n}_i| + |\hat{\omega}_{\theta}\cdot\hat{n}_i|}{2}
\end{equation}
where $\hat{n}_i$ is the unit surface normal of the crack.

Let $(I_r,I_g,I_b)$ be the mean intensity of the light source.
Then the crack intensity, $I_c$ is defined as 
\begin{equation}
    I_c = (I_r,I_g,I_b) \cdot \frac{\mathbb{E}(L_r)}{\sum L_r}
\end{equation}

\subsubsection{Focal Plane and Physical attack simulation}
\label{sec:focal plane}
While we are able to simulate realistic fractures, the primary use case for our work is to be able to generate simulated examples overlayed on existing datasets (KITTI, BDD100k, MS-COCO) and compare them with the real on-road dataset that we created.

Any captured image will exhibit sharp features of the objects in it's focal plane. The glass enclosure covering the camera is extremely close and is thus not part of the focal plane. When the crack happens, the light rays bounce unevenly along the crack and creates a blur (example provided in Fig. \ref{fig:PBR}). We create a binary mask based on the crack pattern and then blur the fractures overlayed on the image. This produces a far-focus image. For a short-focus image, we blur the image and focus on the foreground i.e. the crack.

\section{Experimentation}
\subsection{Dataset}
We benchmark two types of broken glass pattern - real and simulated - on three popular open-source datasets - KITTI \citep{geiger2013IJRR}, BDD100k \cite{yu2020bdd100k} and MS-COCO \citep{lin2014microsoft}. The first two represent specific autonomous driving domain while the last one is a general purpose image dataset. The real broken glass pattern images are collected from FreePik website\footnote{\url{https://www.freepik.com}} and represent the baseline in our case. We collected $65$ images in total and expanded them to a set of $10,000$ images via image augmentation using random shifts, image flips and cropping techniques. We also generate $10,000 $ images using our physics simulator. We then overlay these cracked glass patterns using our PBR pipeline onto every validation image in the datasets and collect the aggregate results. We use three model architectures YOLOv8, Faster R-CNN and PVTv2 model with pretrained weights to generate object detection results.

\subsection{Implementation} 
Our simulation model is developed by randomly sampling $10^4$ particles from a uniform spatial distribution in the given frame in a CPU. A KD-tree from the SciPy python package \citep{2020SciPy-NMeth} using default parameters is constructed to find the approximate nearest neighbors of each particle. A Delaunay triangulation is then run on the particles to create a constrained triangular mesh. We use an impact force of 500 units with a random impact point and a random impact vector. The stress propagation happens until a threshold of 300 units is reached. 
The PBR is performed on CPU by implementing the methods described in the previous section using OpenCV and Python. 

\section{Results and Discussion}
 A major shift from most of the previous works in adversarial examples is that our generated adversarial patterns do not affect all pixels in an image universally. Therefore, the comparison needs to be done only for the image region where the pattern exists. For this purpose, we create a binary mask of each pattern and output the results of the objects which exist in that pattern only. 
\begin{table*}[h!]
    \centering
    \caption{Average precision (in percentage) of different classes in KITTI, BDD100k and MS-COCO under different adversarial images. x provides the overlay relation between dataset and glass-crack type. Clean x Dataset - refers to directly the particular images without any adversarial sample. RO x Dataset - refers to Real images of cracked glass collected online overlayed on clean images. Sim x Dataset - refers to simulated crack patterns overlayed on clean images.}
    \begin{tabular}{c c c c c c}
    \hline
        Dataset & IoU threshold & Category & Clean x Dataset & RO x Dataset & Sim x Dataset\\
        \hline
        \multirow{6}{4em}{KITTI (YOLOv8)} & \multirow{3}{4em}{0.5} & Pedestrian & 25.64 & 69.72 & \textbf{17.84}\\
        & & Truck & 12.39 & \textbf{3.59} & 3.76\\
        & & Car & 58.99 & \textbf{50.7} & 57.73\\
        \cline{2-6}
        &\multirow{3}{4em}{0.75} & Pedestrian & 6.83 & 33.88 & \textbf{6.02}\\
        & & Truck & 11.29 & \textbf{2.67} & 2.79\\
        & & Car & 31.25 & \textbf{23.85} & 30.15\\
        \hline
        \multirow{6}{4em}{\centering BDD100k (PVTv2)}& \multirow{3}{4em}{0.5} & Pedestrian & 66.47 & 54.33 & \textbf{25.95}\\
        & & Truck & 61.97 & \textbf{52.83} & \textbf{52.02}\\
        & & Car & 80.37 & 70.14 & \textbf{56.78}\\
        \cline{2-6}
        & \multirow{3}{4em}{0.75} & Pedestrian & 27.06 & 22.72 & \textbf{10.60}\\
        & & Truck & 47.03 & \textbf{38.23} & 42.52\\
        & & Car & 46.23 & 45.97 & \textbf{42.99}\\
        \hline
        \multirow{6}{4em}{\centering MS-COCO (Faster R-CNN)} & \multirow{3}{4em}{0.5} & Person & 0.035 & 0.024 & \textbf{0.024}\\
        & & Vehicles & 2.14 & \textbf{1.45} & 1.87\\
        & & Food & 35.34 & \textbf{28.07} & 30.65\\
        \cline{2-6}
        & \multirow{3}{4em}{0.75} & Person & 0.032 & \textbf{0.022} & 0.023\\
        & & Vehicles & 1.56 & \textbf{1.05} & 1.07\\
        & & Food & 24.59 & \textbf{18.85} & 22.00\\
        \hline
    \end{tabular}
    \label{tab:benchmark}
\end{table*}

\begin{figure*}[ht]
    \centering
    \includegraphics[width=0.8\textwidth]{images/ablation_elongated_fig.jpg}
    \caption{Top left - Crack pattern collected online on Freepik; top right - online crack pattern overlayed on KITTI; bottom left - simulated crack pattern with PBR; bottom right - simulated crack pattern overlayed on KITTI.}
    \label{fig:data-distributions}
\end{figure*}

Table \ref{tab:benchmark} shows the results of the average precision (AP) under the adversarial images generated using the two types of crack patterns (collected online and simulated) for different classes. For KITTI, the AP of other classes drop as expected with the decrease in AP corresponding to the percentage of image occupied with the truck class recording the highest drop. 
For BDD100K with PVTv2-B0, we see that the drop in AP is largest in the simulated images but overall, the trend is maintained with the pedestrian class showing the steepest drop. 
For MS-COCO, we aggregated the AP for the super - categories: person, vehicles and food. This is because a lot of objects in MS-COCO occupy smaller area in the image frame making it difficult to get meaningful results from all categories.
\begin{figure}[h]
    \centering
    \includegraphics[width=0.45\textwidth]{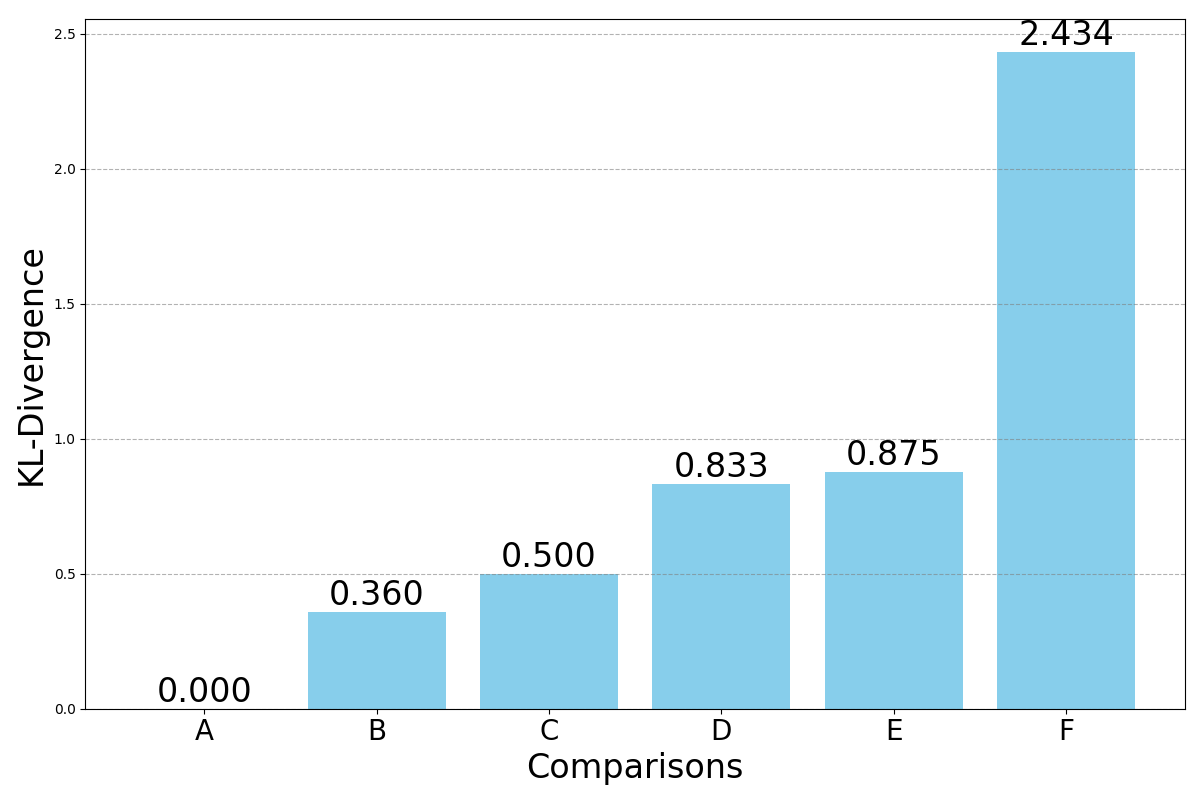}
    \caption{K-L divergence of different pairs of image distributions. Datasets: RC - Real on-road dataset (see Fig. \ref{fig:real_broken_lens}), KITTI and Cats. Filters: RO - Real (collected online) and Sim - Simulated. K-L divergence between (x - overlay relation): A - (Sim x KITTI) vs (Sim x KITTI); B - (Sim vs RO); C - (Clean RC vs KITTI); D - (Broken RC) vs (RO x KITTI); E - (Broken RC) vs (Sim x KITTI); F - KITTI vs Cats.}
    \label{fig:kl}
\end{figure}
A very intriguing result is that the pedestrian class has a multifold increase in AP under the real broken glass patterns. While this trend might seem counter-intuitive, it resonates with the results in Fig. \ref{fig:real_broken_lens} where the confidence of the car increases because of an edge. 
This in fact shows that the AP is highly dependent on the crack pattern making it extremely important to create defense methodologies to mitigate these adversarial attacks.

\subsection{Ablation studies}
Our results indicate that the simulated images obtain a similar adversarial effect as the real images. Thus, an important ablation study for us is to understand how close the simulated crack patterns are to the real cracked glass patterns and those collected online. We form 5 distributions 

\begin{itemize}
    \item Real on-road dataset (depicted in Fig. \ref{fig:real_broken_lens})
    \item Crack patterns collected online (Fig. \ref{fig:data-distributions} top left)
    \item Simulated crack patterns (Fig. \ref{fig:data-distributions} bottom left)
    \item Simulated crack patterns overlayed on KITTI (Fig. \ref{fig:data-distributions} bottom right)
    \item Crack patterns collected online overlayed on KITTI (Fig. \ref{fig:data-distributions} top right)
\end{itemize}
We now compute the K-L divergence among all these distributions to compute how similar they are to each other (see Fig. \ref{fig:kl}). In order to provide a control, we compare KITTI to images of cats from the Kaggle dataset, providing a K-L divergence of 2.434. In that scale, the PBR images of broken glass have a difference of 0.36 to the real broken glass patterns while the broken glass filters overlaid on KITTI images have similar K-L divergence.

\begin{figure}[h]
    \centering
    \includegraphics[width=1.0\linewidth]{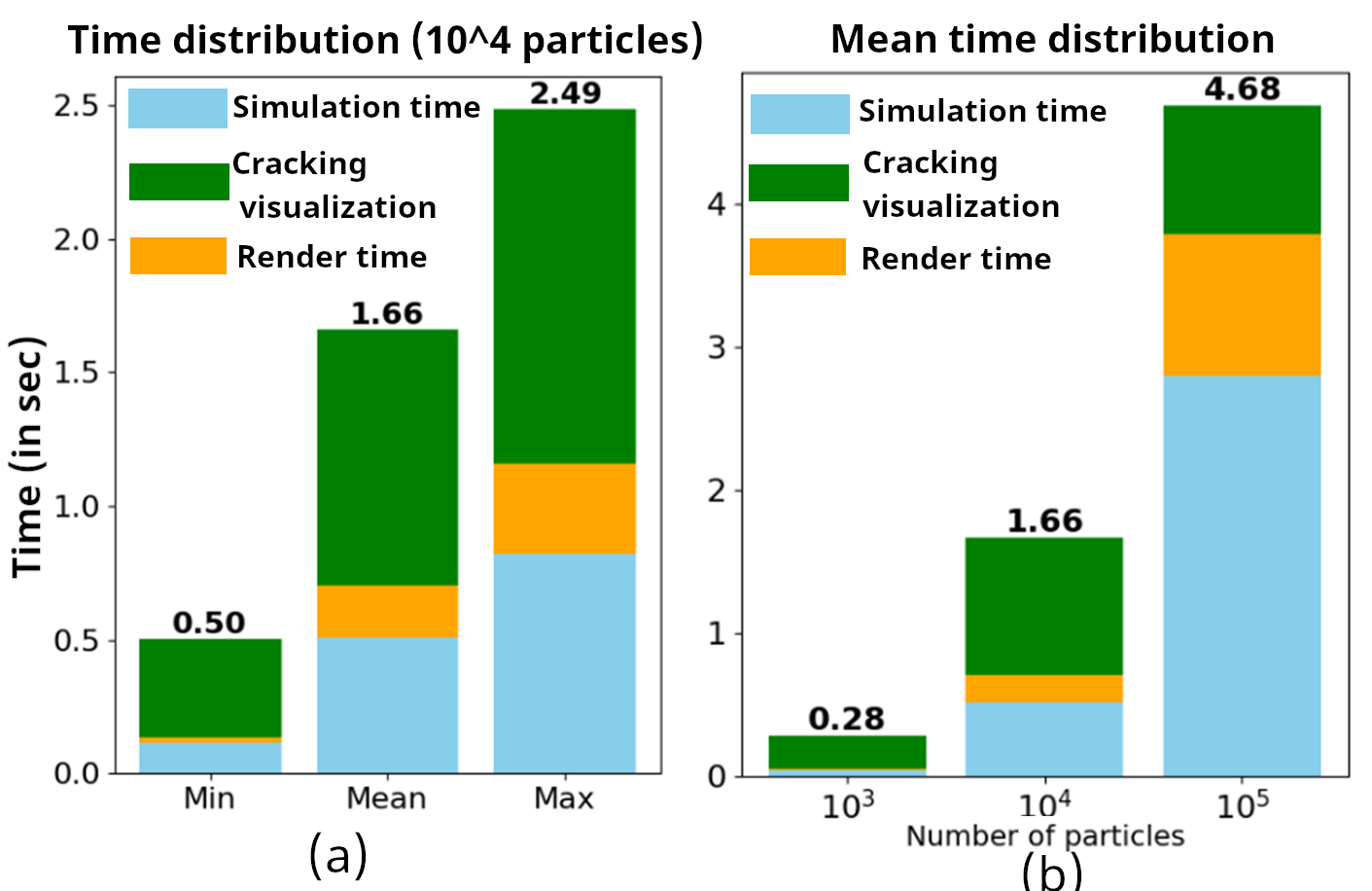}
    \caption{(a) Mean time taken by different modules of our pipeline across 100 runs. (b) The minimum, maximum and mean time taken by different modules across 100 runs for a $10^4$ particle mesh. For these plots, we showcase the time taken for simulation (simulation time), converting the mesh to glass (cracking visualization) and finally rendering (render time).}
    \label{fig:time-distribution}
\end{figure}

Fig. \ref{fig:time-distribution} shows an analysis of the computation time for each of our modules and over different number of particles. We perform this analysis on 100 runs, generating random impact points, impact angles, and mesh structure with a fixed number of particles. The difference in computation time for different runs can be attributed to the impact point and impact angle. 
The cracking visualization and render time also vary owing to different sized masks formed due to varying fracture patterns. We also vary the number of particles and see how runtime increases exponentially with the increase in particles. All these runs were rendered on images from the KITTI dataset with dimensions of $(375 \times 1242 \times 3)$.

\section{Conclusion and Future Scope}
We have introduced a novel class of adversarial failures resulting from the physical process of failures in the camera. 
In this paper, we provide an approach to generate a realistic broken glass pattern from a physical simulation and subsequently embed that to existing image datasets using physically based rendering. We show that the simulated adversarial images can lead to significant errors in object detection. 

In this work, we address black-box adversarial attacks stemming from real-world, naturally occurring physical phenomena, not artificially crafted to exploit specific model vulnerabilities. We assume no knowledge of the model attributes, weights or architecture, ensuring attacks are transferable across various models. Physical adversarial methods (Translucent Patch, RP2) can all be termed as occlusions of either the camera or the objects being captured. The adversariality comes from the effect of the model inference due to these occlusions. Our PBR pipeline blends the cracks with source images as translucent, blurry patterns, impacting latent space encoding rather than causing direct occlusion, resulting in incorrect detections. 

While this work introduces a physics-based method for broken glass pattern generation specifically, camera failures encompass other effects such as sun-glare, overexposure, underexposure, condensation etc. Our future work will focus on creating an adversarial toolbox for realistic generation of these  effects using physics and subsequently, placing them on existing image datasets and car simulation platforms to promote further research in this field of partial camera failures.

\bibliography{aaai2026}

\clearpage

\appendix
\section{Algorithm of stress propagation}
Algorithm~\ref{alg:sim_alg} describes the procedure for simulating the propagation of stress through a material following an impact event. The algorithm takes as inputs the location of the impact ($pt$), the magnitude of the impact force ($F$), the impact direction vector ($v$), and the parent edge ($PE$) associated with the impact site. It also uses a nearest neighbor radius $R$ to determine the set of candidate locations for stress propagation.

\begin{algorithm}[h!]
    \caption{Stress Propagation}
    \label{alg:sim_alg}
    \begin{algorithmic}[1]
        \State $pt\leftarrow \text{Impact Point}$
        \State $F\leftarrow \text{Impact Force}$
        \State $PE\leftarrow \text{Parent Edge}$
        \State $v\leftarrow \text{Impact Vector}$
        \State $R \leftarrow \text{Nearest neighbor radius}$
        \State
        \Procedure{PropagateStress}{$Pt,F,V,PE$}
            \State $frontiers \leftarrow KDTree-queryRadius(R)$
            \State $NN \leftarrow \frac{frontiers - pt}{||frontiers - pt||}$
            \State $cos(\theta) \leftarrow NN \cdot v$
            \State $stress\leftarrow calculateStress(cos(\theta),F)$
            \State $frontiers \leftarrow frontiers[argmax(stress)]$
            \State $v \leftarrow v[argmax[stress]]$
            \State $PE \leftarrow PE[argmax[stress]]$
            \State \Call{PropagateStress}{$Pt,F,V,PE$}
        \EndProcedure
    \end{algorithmic}
\end{algorithm}
First, it uses a KD-tree data structure to efficiently query all points (frontiers) within a given radius $R$ of the impact point. For each frontier, it computes a unit direction vector from the impact point to the frontier ($NN$). It then projects the impact vector $v$ onto this direction to obtain the cosine similarity $cos(\theta)$, capturing the angular relationship between the impact direction and the candidate propagation direction. For each candidate, the resulting value is used, together with the impact force, to calculate the corresponding stress at that point. The algorithm then selects the candidate with the maximum stress value. The impact vector $v$ and parent edge $PE$ are updated to correspond to this new direction. The process is recursively repeated, allowing the simulated stress wave to propagate iteratively through the material along the path of greatest stress transfer.

This approach aims to mimic how stress from an impact point is most likely to radiate through a material—preferentially following paths defined by both geometric proximity and mechanical alignment with the original impact.

The final output of the simulation is the realization of the mesh as an image which corresponds to broken lens pattern (final image of Fig. \ref{fig:crack-animation}). 
\begin{figure*}[!htbp]
    \centering
    \includegraphics[width=\textwidth]{images/glass_animation.png}
    \caption{An animation of fracturing of a lens simulated by setting the stress field and applying PBR.}
    \label{fig:crack-animation}
\end{figure*}

\clearpage

\section{Static Experiment}
\label{sec:static}
In order to understand the effect of these fractures on the resultant images, we first conduct a indoor static experiment as referenced in Section \nameref{sec:intro}. We use various tempered glass sheets for this experiment, which we break randomly using a small hammer with one single or multiple break points. Then, we place a 36 MP JVC GC-PX10 hybrid camera mounted on a tripod with a clamp in front for the tempered glass. 

\begin{figure}[b]
    \centering
    \includegraphics[width=0.4\textwidth]{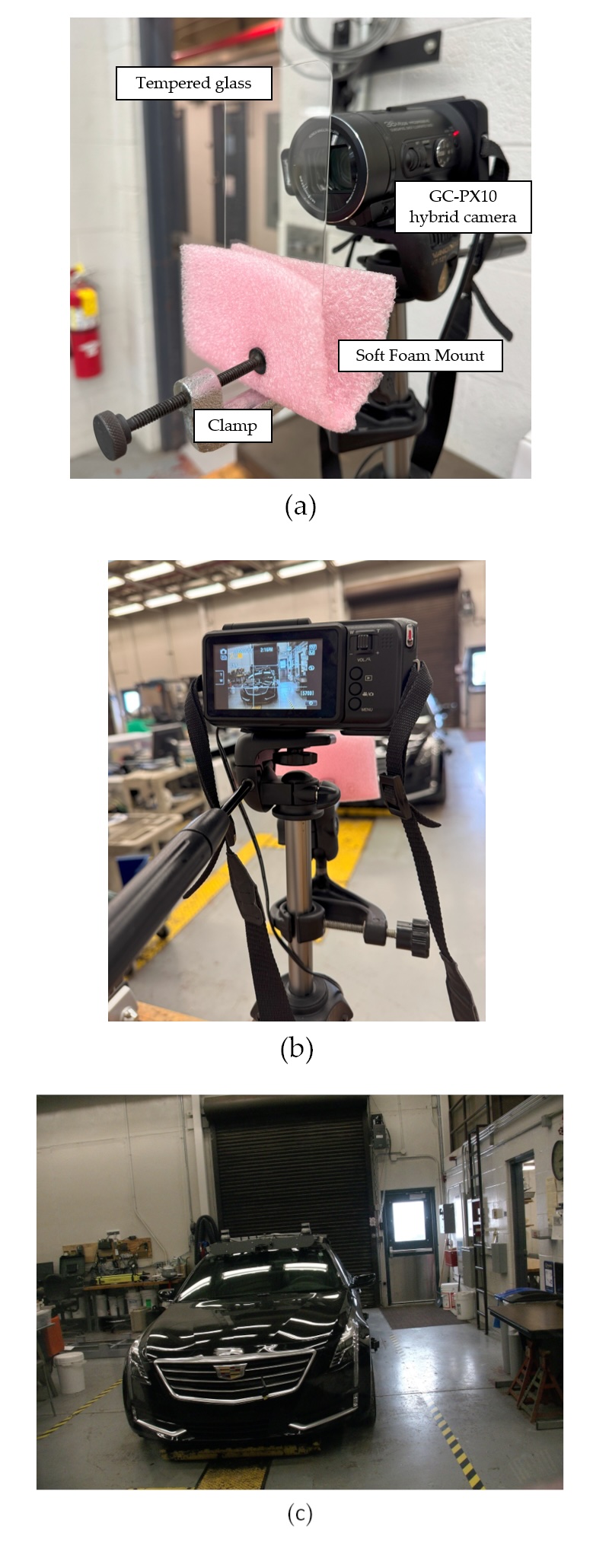}
    \caption{Experimental setup for collecting images impacted by scratched/broken outer layers for a camera. (a) shows the entire setup for taking adversarial images. (b) shows the position of the camera w.r.t. the scene being captured. (c) shows the scene being captured by the camera}
    \label{fig:supp_experimental_setup}
\end{figure}

Fig.~\ref{fig:supp_experimental_setup}(a) shows the detailed setup with the camera mount and tempered glass held in place with a clamp. Fig.~\ref{fig:supp_experimental_setup}(b) shows the image captured by the camera and the Fig.~\ref{fig:supp_experimental_setup}(c) shows the single vehicle placed as the primary object being captured by the camera through the tempered glass. The scene is illuminated using overhead fluorescent lights. 

\begin{figure*}[ht]
    \centering
    \includegraphics[width=0.8\textwidth]{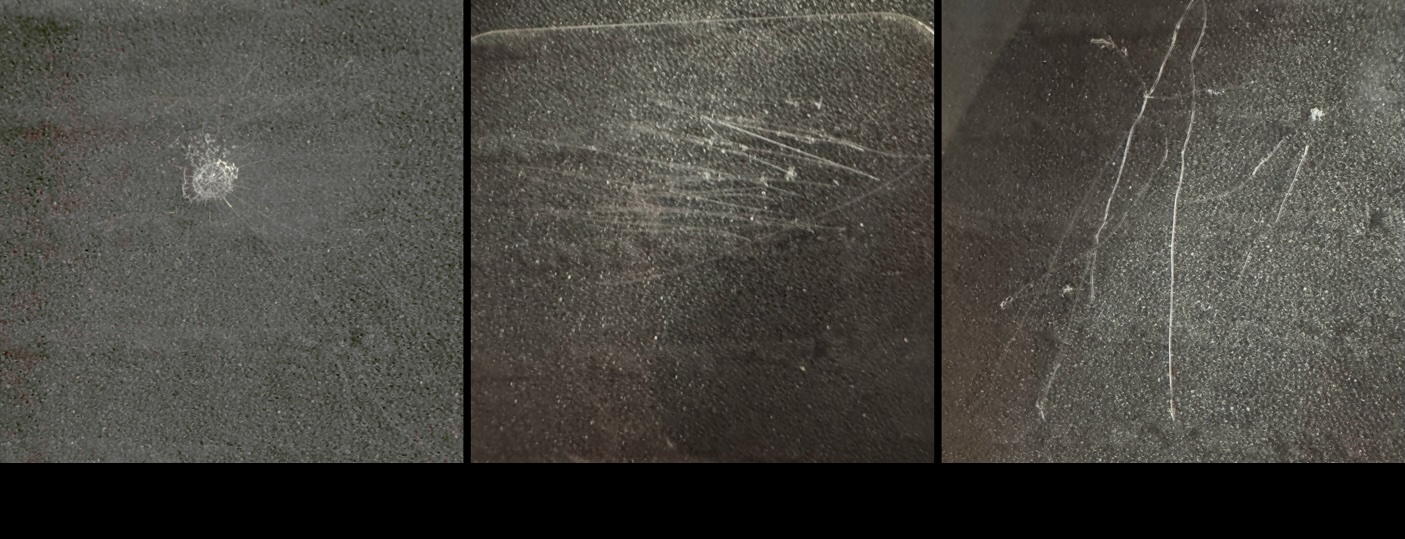}
    \caption{Some fractures/scratched patterns on the glass we used for collecting the images. (a) A sharp force applied perpendicular to the glass surface, producing fractures occurring radially. (b) and (c) replicate a glass with scratches}
    \label{fig:supp_broken_glasses}
\end{figure*}

 Fig.~\ref{fig:supp_broken_glasses} shows some of the fractures/scratched patterns on the tempered glass. These patterns were intentionally randomized, employing multiple focal points and different levels of force to mimic the unpredictable and varied nature of real-world glass damage. By applying diverse force strengths, we were able to produce a spectrum of fractures and scratches, ranging from fine surface abrasions to more pronounced fractures. This approach was chosen to closely replicate the types of damage that glass surfaces may encounter in actual conditions—such as those caused by impacts, debris, or environmental stressors—thereby ensuring the relevance and realism of our experimental setup. These representative damage patterns allow us to more effectively analyze the influence of glass imperfections on sensor performance and object detection algorithms.
 
\begin{figure*}[ht]
    \centering
    \includegraphics[width=\textwidth]{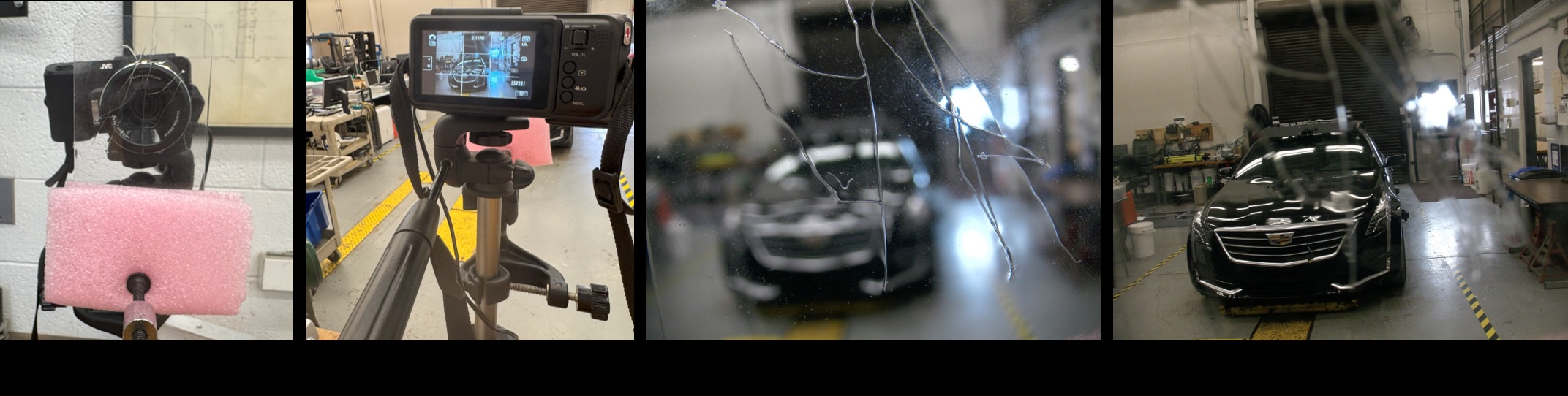}
    \caption{(a) Shows the scratched pattern placed in front of the camera, (b) shows the camera POV. (c) shows the image captured by the camera (short-focus). (d) shows the image captured by the camera (far-focus)}
    \label{fig:supp_exp1}
\end{figure*}

\begin{figure*}[ht]
    \centering
    \includegraphics[width=\textwidth]{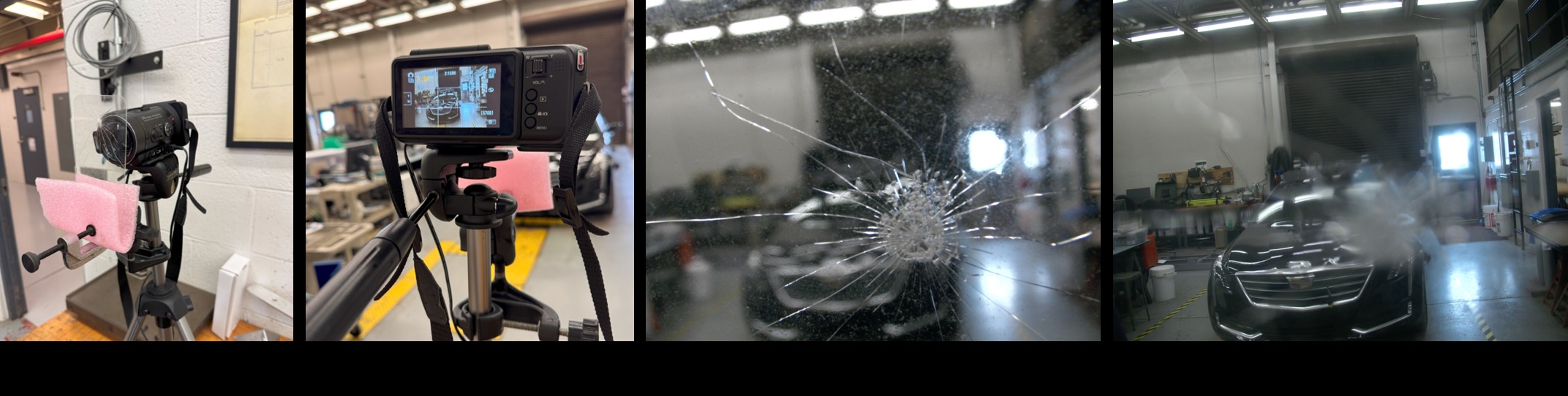}
    \caption{(a) Shows the broken glass pattern in front of the camera, (b) shows the camera POV. (c) shows the image captured by the camera (short-focus). (d) shows the image captured by the camera (far-focus)}
    \label{fig:supp_exp2}
\end{figure*}

Two different fracture patterns and their resultant images are shown in Fig.~\ref{fig:supp_exp1}
and Fig.~\ref{fig:supp_exp2}. We would like to note that we varied the focal lengths of the camera considerably to understand how the images look under near- and far-focus. The outputs show that even minor scratched patterns show up in the image output whereas much stronger multi-fracture pattern can blur almost the entire image. This experiment provides the intuition on which our simulation and visualization framework is built. 

\begin{figure}
    \centering
    \includegraphics[width=\linewidth]{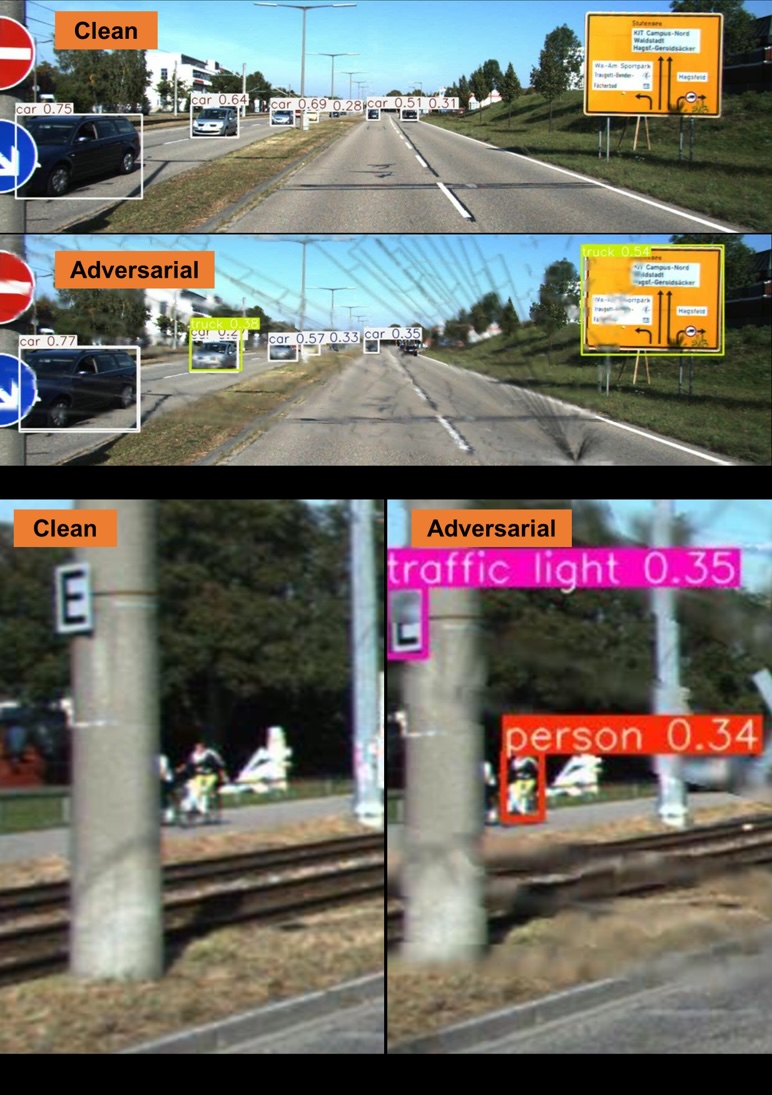}
    \caption{(a) Top - detections on a clean image; bottom - detections on an adversarial image.(b) YoLo fails to detect the person (c) Glass cracks allows the model to detect the person.}
    \label{fig:false-detections}
\end{figure}
\section{Increased AP for pedestrians in KITTI}
We would like to point out that the increased AP for the pedestrian class was something that even we were surprised at first. However, a careful-qualitative deep-dive analysis helped us understand that this was occurring as a result of the glass cracks making it easier for the model to classify pedestrians because of enhanced edges around them. This wasn’t an edge artifact but instead the glass crack acting as an additional edge boundary clearly separating the pedestrian and the background. A similar result was also observed in [1] where the overall AP was increased in adversarial images.
\clearpage

\section{Dynamic Experiment}
\label{sec:dynamic}
\begin{figure*}
    \centering
    \includegraphics[width=0.8\textwidth]{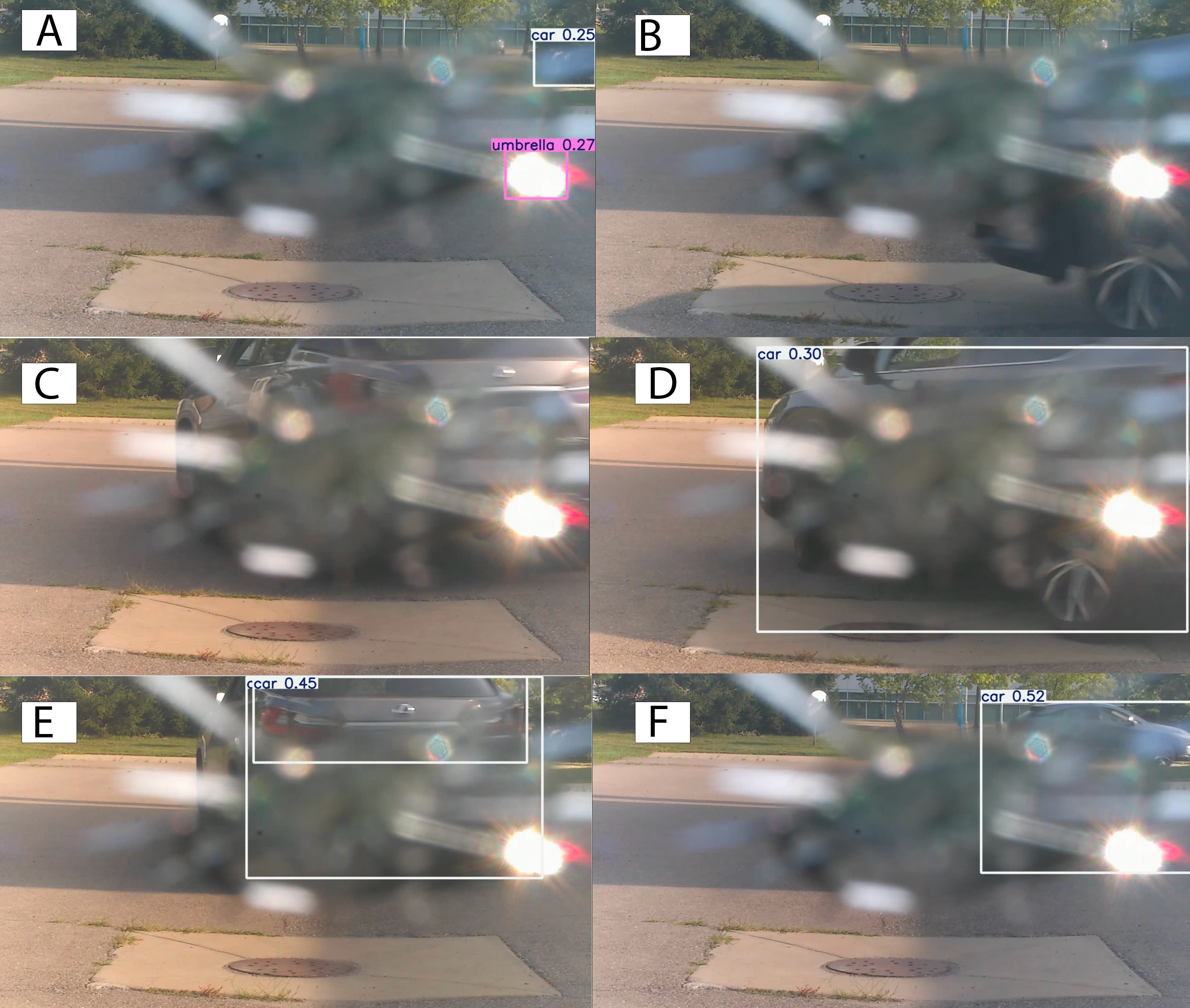}
    \caption{Specific frames of the images taken with the windshield crack with YOLOv8 inference for the vehicle class. A - false positive with no object in scene; B - no inference on vehicle; C - no inference on vehicle; D - first detection on vehicle; E - two different detections on the same vehicle; F - wrong bounding box area.}
    \label{fig:dynamic-vehicle}
\end{figure*}

\begin{figure*}
    \centering
    \includegraphics[width=0.8\textwidth]{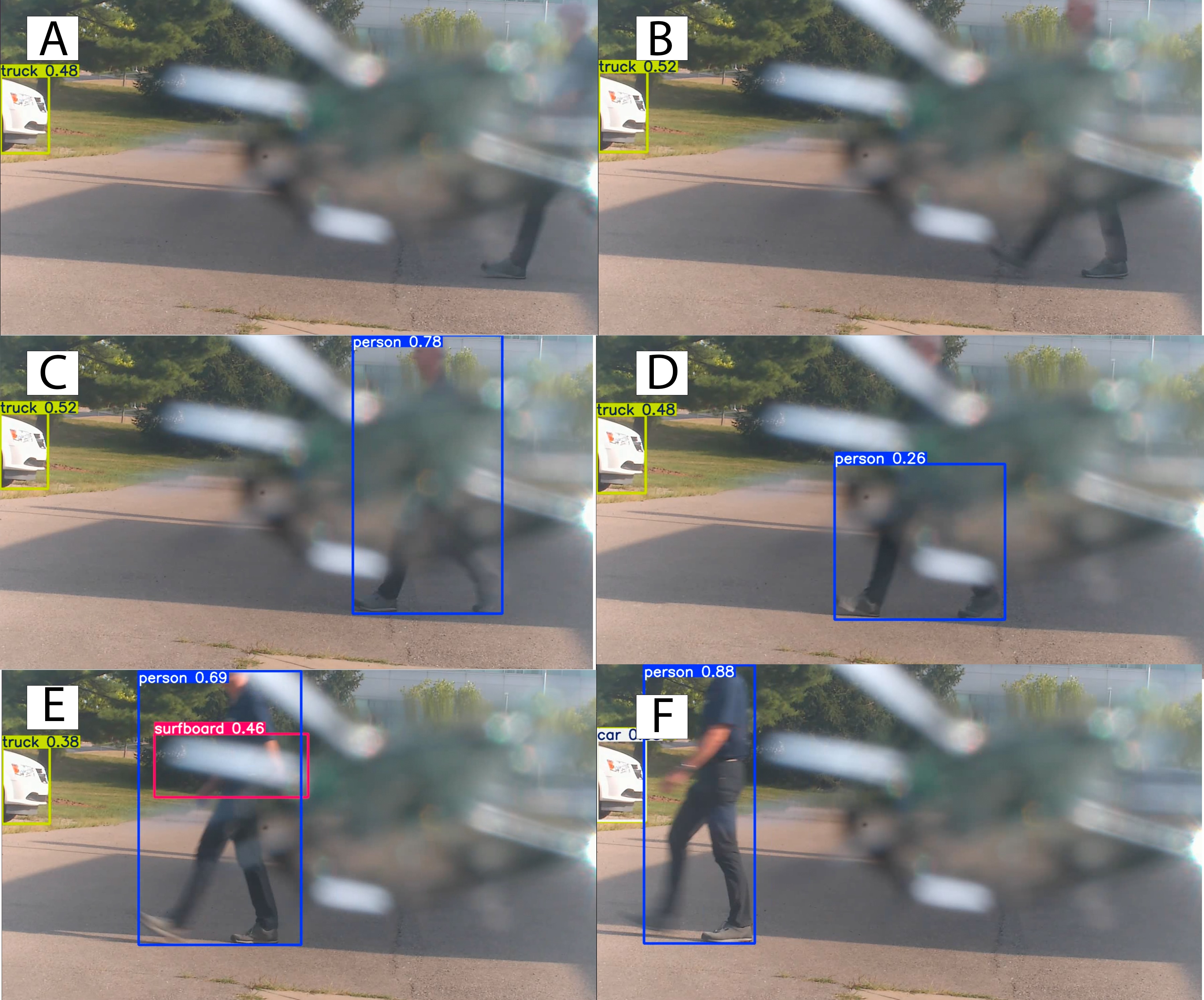}
    \caption{Specific frames of the images taken with the windshield crack with YOLOv8 inference for the person class. A - first entry of person in scene with no detection; B - no inference of person; C - first detection of person; D - partial detection of person; E - detection of person with other class; F - full detection of person.}
    \label{fig:dynamic-person}
\end{figure*}

In this section, we describe the dynamic experiment mentioned in Section \nameref{sec:intro}. We perform this experiment to understand the temporal perturbation introduced by a crack. We use a windshield crack of a vehicle and place a small camera on the dashboard behind the crack. Then we photograph two dynamic objects - a vehicle and a pedestrian as they move across the scene. Fig. \ref{fig:dynamic-vehicle} provides some specific image frames with inference from YOLOv8 for the vehicle class. We show that with the crack, the vehicle remains undetected in most frames. Additionally, almost every frame contains a false positive. Correspondingly, we present Fig. \ref{fig:dynamic-person} as the frames with a person walking in the scene. We show that it intermittently provides detection and occasionally with a wrong class (surfboard). 

\clearpage

\clearpage

\section{Real glass fracture images}
\label{sec:real_glass_fracture}
\begin{figure*}
    \centering
    \includegraphics[width=\textwidth]{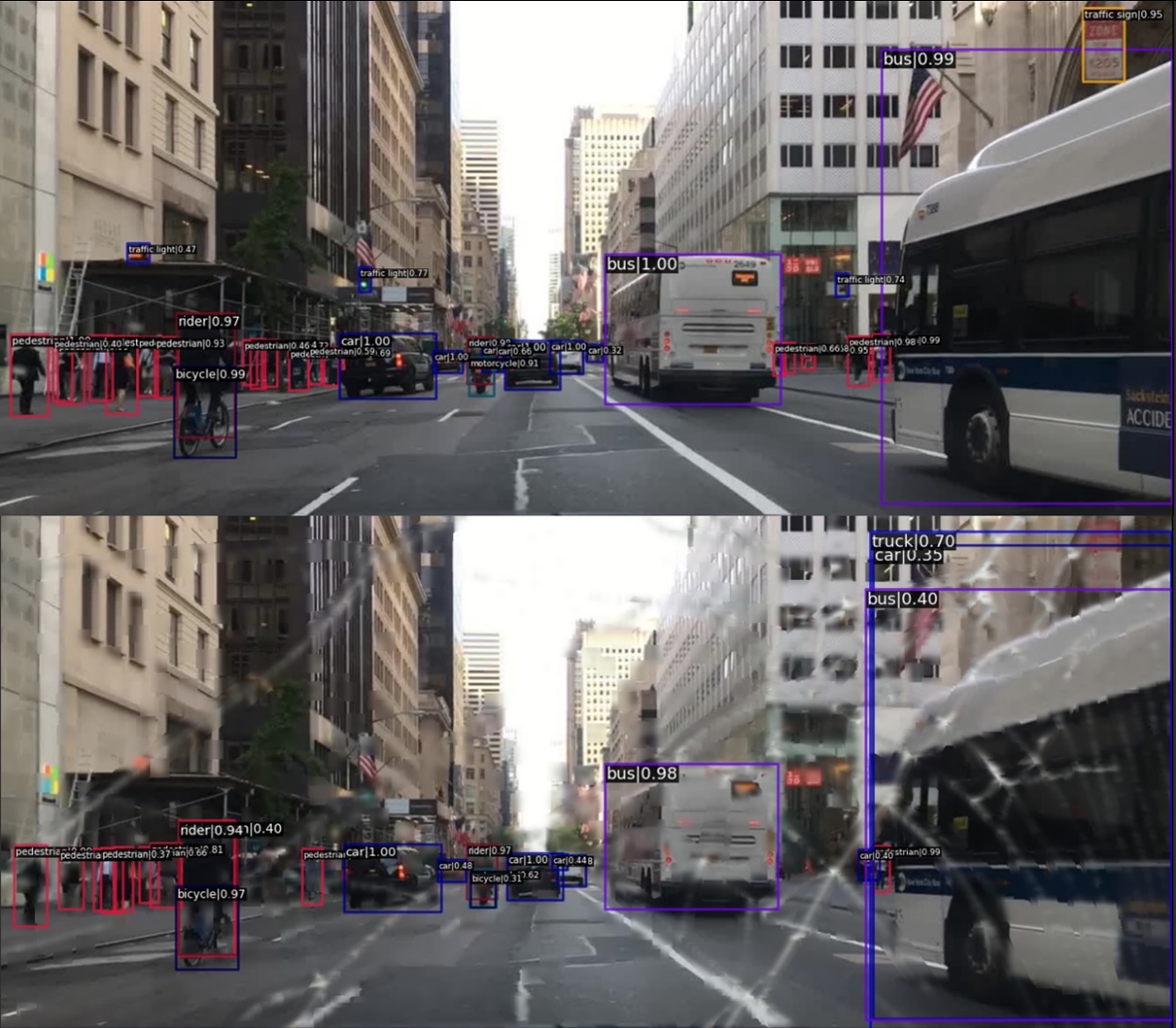}
    \caption{Top - Inference of PTv2 on a clean image from BDD100k. Bottom - Inference for a real broken glass image overlaid on BDD100k for comparison. We see two extra false positives in on the right side (truck, car) and several false negatives for the pedestrian class on the left of the adversarial image.}
    \label{fig:kitti_glass_fracture}
\end{figure*}

We present an example of the glass fracture images collected from the FreePik website overlaid on KITTI dataset along with YOLOv8 inference (Fig. \ref{fig:kitti_glass_fracture}). We show that the fracture removes some detections and decreases the detection confidence of others. 

\end{document}